\RequirePackage{snapshot}
\documentclass[10pt,twocolumn,letterpaper]{article}

\usepackage{cvpr}
\usepackage{times}
\usepackage{epsfig}
\usepackage{graphicx}
\usepackage{amsmath}
\usepackage{amssymb}

\usepackage{booktabs}
\usepackage{tabularx}
\usepackage[font=small]{caption}
\usepackage{subcaption}
\usepackage{tablefootnote}
\usepackage{adjustbox}
\usepackage{textcomp}  
\usepackage{xcolor,colortbl}  
\usepackage[para]{footmisc}  
\usepackage[square,numbers,sort&compress]{natbib}
\usepackage{color}
\usepackage[normalem]{ulem}

\newcommand{\tx}[0]{Action Transformer}
\newcommand{\Tx}[0]{Action Transformer}
\newcommand{\QPr}[0]{QPr}  
\newcommand{\qprconcat}[0]{HighRes}
\newcommand{\qpravg}[0]{LowRes}

\newcommand{\st}[0]{spatiotemporal }

\newcommand{\tableSize}[0]{\footnotesize}

\renewcommand{\footnotesize}{\scriptsize}

\definecolor{Gray}{gray}{0.9}
\newcolumntype{g}{>{\columncolor{Gray}}c}  
\newcolumntype{H}{>{\setbox0=\hbox\bgroup}c<{\egroup}@{}}  
\definecolor{GrayLine}{gray}{0.7}
\newcolumntype{Y}{>{\centering\arraybackslash}X}
\DeclareMathOperator*{\Softmax}{Softmax}

\usepackage[pagebackref=true,breaklinks=true,letterpaper=true,colorlinks,bookmarks=false]{hyperref}

\cvprfinalcopy 

\def\cvprPaperID{292}

\ifcvprfinal\fi 
\begin{document}

\title{Video Action Transformer Network}

\author{
Rohit Girdhar$^{1\thanks{Work done during an internship at DeepMind}}$ \quad
Jo{\~a}o Carreira$^{2}$ \quad
Carl Doersch$^{2}$ \quad
Andrew Zisserman$^{2,3}$ \\
$^{1}$Carnegie Mellon University \quad $^{2}$DeepMind \quad $^{3}$University of Oxford \\
{\small \url{http://rohitgirdhar.github.io/ActionTransformer}}
} 
\maketitle

\begin{abstract}
We introduce the \Tx{} model for 
recognizing and localizing human actions in video clips. 
We repurpose a Transformer-style architecture to aggregate features from the spatiotemporal context around the person whose actions we are trying to classify.
We show that by using  high-resolution, person-specific,  class-agnostic queries, the model spontaneously learns to track individual people and to pick up on semantic context from the actions of
others. Additionally its attention mechanism learns to emphasize hands and faces, which are often crucial to
discriminate an action -- all without explicit supervision other than boxes and class labels.
We train and test our \Tx{} network on the Atomic Visual Actions
(AVA) 
dataset, outperforming the state-of-the-art  by a significant
margin 
using 
only raw RGB frames as input.

\end{abstract} \section{Introduction}

In this paper, our objective is to both localize and recognize human actions in video clips. 
One reason that human actions remain so difficult to recognize
is that inferring a person's actions often requires understanding the
people and objects around them.  
For instance, recognizing whether a person is `listening to someone'
is predicated on the existence of another person in the scene 
saying something.  Similarly, recognizing whether a person is
`pointing to an object', or `holding an object', or `shaking hands';
all require reasoning jointly about the person and the
animate and inanimate elements of their surroundings. Note that this
is not limited to the context at a given point in time: recognizing
the action of `watching a person', after the watched person has walked out of
frame,  requires reasoning over time to understand that our
person of interest is actually looking at someone and not just staring
into the distance.

\begin{figure}
\centering
\includegraphics[width=\linewidth]{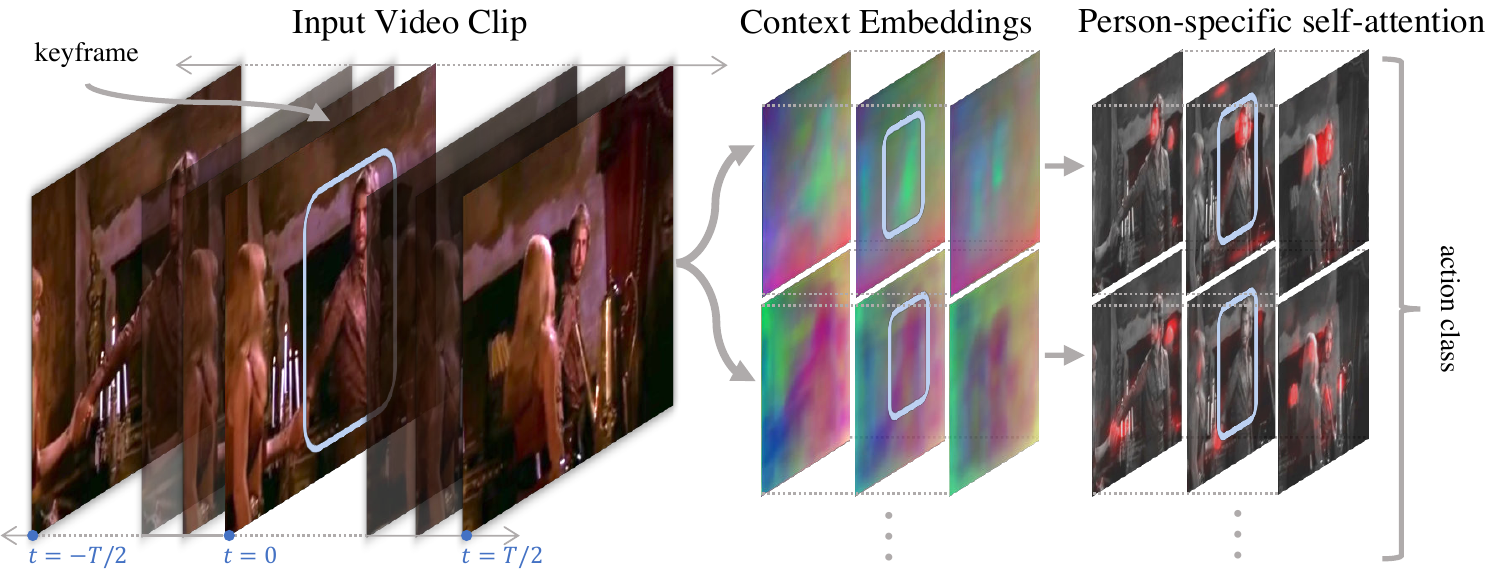}
\caption{
{\bf \Tx{} in action.}
Our proposed multi-head/layer \Tx{} architecture learns to attend to relevant regions of the person of interest and their context (other people, objects) to recognize the actions they are doing.
Each head computes a clip embedding, which is used to focus on different parts like the face, hands and the other people to recognize that the person of interest is `holding hands' and `watching a person'.
}\label{fig:teaser}
\end{figure}

Thus we seek a model that can determine and utilize such contextual
information (other people, other objects) when determining the action
of a person of interest. The Transformer architecture from Vaswani
{\it et al.}~\cite{vaswani2017attention} is one suitable model for
this, since it explicitly builds contextual support for its
representations using self-attention.  This architecture has been
hugely successful for sequence modelling tasks
compared to traditional recurrent models.
The question, however, is: how does one build a similar model for human action
recognition?

Our answer is a new video action recognition network, the {\bf \Tx{}}, that
uses a modified Transformer architecture as a `head' to classify the action of
a person of interest. It brings together two other ideas: (i) a
spatio-temporal I3D model that has been successful in previous
approaches for action recognition in video~\cite{carreira2017quo} --
this provides the base features; and (ii) a region proposal network
(RPN)~\cite{ren2015faster} -- this provides a sampling mechanism for localizing people performing actions. Together the I3D features
and RPN generate the query that is the input for the
Transformer head that
aggregates contextual information from other people and objects in the
surrounding video. We describe this architecture in detail in section~\ref{sec:app}.
We show in section~\ref{sec:exp} that the trained network 
 is able to learn both to track individual people and to contextualize their actions in terms of the actions of other people in the video. In addition, the transformer attends to hand and face regions, which is reassuring because we know they have some of the most relevant features
 when discriminating an action. All of this is obtained without explicit supervision, but is instead
learned during action classification.

We train and test our model on the Atomic Visual Actions
(AVA)~\cite{gu2018ava} dataset. This
is an interesting and suitable testbed for this kind of contextual reasoning. It requires detecting multiple people in videos semi-densely in time, and recognizing multiple basic actions. Many of these actions often cannot be determined from the person bounding box alone, but instead require inferring relations
to other people and objects. Unlike previous works~\cite{baradel2018object},
our model learns to do so without needing explicit object detections.
We set a new record on the 
AVA dataset, improving performance from 17.4\%~\cite{sun2018arcn} to 25.0\% mAP. The network 
only uses raw RGB frames, yet it
outperforms all previous work, including large ensembles that use additional optical flow and sound inputs. At the time of submission, ours was the top performing approach on the ActivityNet leaderboard~\cite{ava_leaderboard}. 

However, we note that at 25\% mAP, this problem, or even this dataset, is far from solved.
Hence, we rigorously analyze the failure cases of our model in Section~\ref{sec:analysis}. We 
describe some common failure modes and analyze the performance broken down by semantic and spatial labels. 
Interestingly, we find many classes with relatively large training sets are still hard to recognize. 
We investigate such tail cases to flag potential avenues for future work.

 \section{Related Work}

{\noindent \bf Video Understanding:} Video activity recognition has
evolved rapidly in recent years. Datasets have become progressively
larger and harder: from actors performing simple
actions~\cite{gorelick2007weizmann,schuldt2004kth}, to short sports
and movie clips~\cite{ucf101,hmdb51}, finally to diverse YouTube
videos~\cite{kay2017kinetics,youtube8M}. Models have followed suit, from
hand-crafted features~\cite{laptev2005space,IDT_Wang_13} to deep
end-to-end trainable
models~\cite{Karpathy_14,WangL_16a,carreira2017quo,xie2017rethinking,wang2017non}.
However, much of this work has focused on trimmed action recognition,
i.e., classifying a short clip into action classes. While useful, this
is a rather limited view of action understanding, as most videos
involve multiple people  performing multiple different actions at any
given time.  Some recent work has looked at such fine-grained video
understanding~\cite{singh2017online,hou2017tube,duarte2018videocapsule,kalogeiton2017action},
but has largely been limited to small datasets like
UCF-24~\cite{ucf101,singh2017online} or JHMDB~\cite{Jhuang2013JHMDB}.
Another thread of work has focused on temporal action
detection~\cite{charades,sigurdsson2017asynchronous,xu2017rc3d};
however, it does not tackle the tasks of person detection or person-action
attribution.

{\noindent \bf AVA dataset and methods:} The recently introduced AVA~\cite{gu2018ava} dataset has attempted to remedy this by introducing 15-minute long clips labeled with all people and their actions at one second intervals. Although fairly new, various models~\cite{gu2018ava,sun2018arcn,tsinghua_ava,yh_ava_submit} have already been proposed for this task. Most models have attempted to extend object detection frameworks~\cite{he2017mask,ren2015faster,huang2017speed} to operate on videos~\cite{hou2017tube,kalogeiton2017action,girdhar2018detecttrack}. Perhaps the closest to our approach is the concurrent work on person-centric relation networks~\cite{sun2018arcn}, which learns to relate person features with the video clip akin to relation networks~\cite{santoro2017simple}. In contrast, we propose to use person
detections  as queries to seek out regions to aggregate in order to recognize their actions, and outperform~\cite{sun2018arcn} and other prior works by a large margin.

{\noindent \bf Attention for action recognition:}
There has been a large body of work on incorporating attention in neural networks, primarily focused on language related tasks~\cite{vaswani2017attention,xu2015show}. Attention for videos has been pursued in various forms, including gating or second order pooling~\cite{xie2017rethinking,Girdhar_17b_AttentionalPoolingAction,long2018attention,miech17loupe}, guided by human pose or other primitives~\cite{Baradel_2018_BMVC,Girdhar_17a_ActionVLAD,baradel2017human,Girdhar_17b_AttentionalPoolingAction}, region-graph representations~\cite{herzig2018classifying,wang2018spacetime}, recurrent models~\cite{sharma2015attention} and self-attention~\cite{wang2017non}. Our model can be thought of as a form of self-attention complementary to these approaches.
Instead of comparing all pairs of pixels, it reduces one side of the comparison to human regions, 
and can be applied on top of a variety of base architectures, including the previously mentioned attentional architectures like~\cite{wang2017non}.
 \begin{figure*}[t]
\centering
\includegraphics[width=\linewidth]{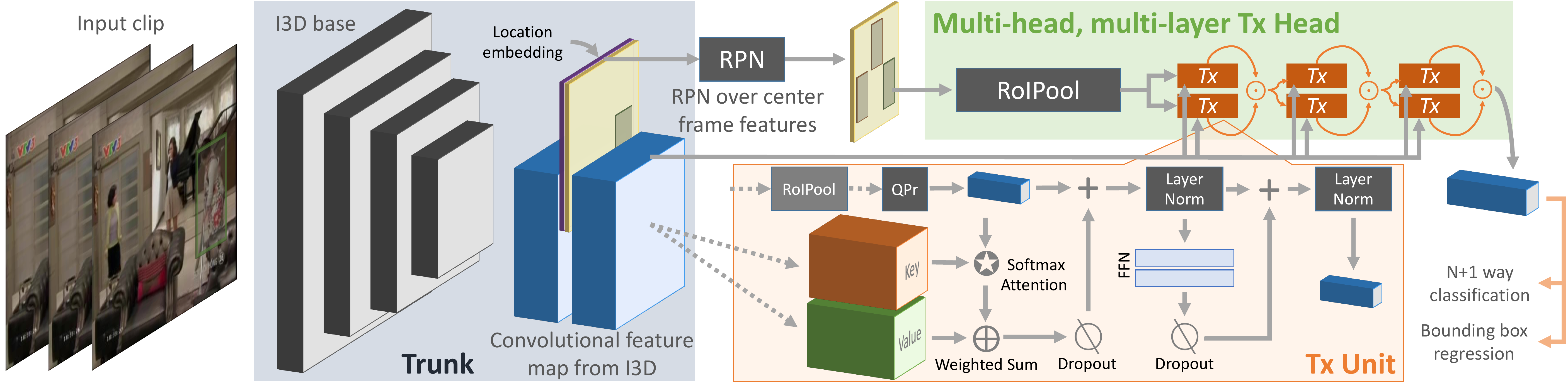}
\caption{
{\bf Base Network Architecture.}
Our model takes a clip as input and generates a spatio-temporal feature representation using a trunk network, typically the initial layers of I3D. The center frame of the feature map is passed through an RPN to generate bounding box proposals, and the feature map (padded with location embedding) and each proposal are passed through `head' networks to obtain a feature for the proposal. This feature is then used to regress a tight bounding box and classify into action classes.
The head network consists of a stack of \tx{} (Tx) units, which generates the features to be classified. We also visualize the Tx unit zoomed in, as described in Section~\ref{sec:app:tx_head}.
\QPr{} and FFN refer to Query Preprocessor and a Feed-forward Network respectively, also explained Section~\ref{sec:app:tx_head}.
}\label{fig:base_arch}
\end{figure*}

\section{\Tx{}  Network}\label{sec:app}

In this section we describe the overall design of our new \Tx{} model.
It is designed  to detect all persons, and
classify all the actions they are doing, at a given time point
(`keyframe').  It ingests a short video clip 
centered on the keyframe, and generates
a set of human bounding boxes for all the people in the central frame,
with each box labelled with all the
predicted actions for the person.

The model  consists of a distinct base and head networks, similar to the
Faster R-CNN object
detection framework~\cite{ren2015faster}.  The base, which we also refer to as trunk, uses a 
3D convolutional
architecture to generate features and region
proposals (RP) for the people present. The head then uses the features associated with each proposal to
predict actions and regresses a tighter bounding box. Note that, importantly, both the RPN and bounding box
regression 
are action agnostic. 
In more detail, the head 
uses the feature map generated by the
trunk, along with the RPN proposals, to generate a feature
representation corresponding to each RP using RoIPool~\cite{huang2017speed} operations.
This feature is then used classify the box into $C$ action classes or background (total $C+1$),
and regress to a 4D vector of offsets to convert the RPN proposal into
a tight bounding box around the person.  The base  is described in
Section~\ref{sec:app:base_arch}, and the transformer head in
Section~\ref{sec:app:tx_head}. We also describe an alternative
I3D Head in
Section~\ref{sec:app:i3d_head}, which is a more direct analogue of the
Faster-RCNN head. It is used in the ablation study.  Implementation
details are given in Section~\ref{sec:app:impl}.

\subsection{Base network architecture}\label{sec:app:base_arch}

We start by
extracting a $T$-frame (typically 64) clip from the original video,
encoding about 3 seconds of context around a given keyframe. We encode
this input using a set of convolutional layers, and refer to this
network as the trunk. In practice, we use the initial layers of an I3D
network pre-trained on Kinetics-400~\cite{carreira2017quo}. We extract
the feature map from the {\tt Mixed\_4f} layer, by which the $T\times
H\times W$ input is downsampled to $T'\times H'\times W' = \frac{T}{4}
\times \frac{H}{16} \times \frac{W}{16}$. We slice out the temporally-central 
frame from this feature map and pass it through a region
proposal network (RPN)~\cite{ren2015faster}.  The RPN generates
multiple potential person bounding boxes along with objectness
scores. We then select $R$ boxes (we use $R=300$) with the highest objectness scores to be
further regressed into a tight bounding box and classified into the
action classes using a `head' network, as we describe next.  The trunk
and RPN portions of Figure~\ref{fig:base_arch} illustrate the network
described so far.

\subsection{\Tx{} Head}\label{sec:app:tx_head}

As outlined  in the Introduction, our head architecture 
is inspired and
re-purposed from the Transformer
architecture~\cite{vaswani2017attention}. It uses the person box from the RPN as a
`query' to locate regions to attend to, and aggregates the information
over the clip to classify their actions.  We first 
briefly review  the Transformer architecture, and then describe our
\Tx{} head framework.

{\noindent \bf Transformer:}  This architecture was
proposed in~\cite{vaswani2017attention} for
seq2seq tasks like language translation, to replace traditional
recurrent models. The main idea of the original architecture is to compute
{\em self-attention} by comparing a feature to all other features in the sequence.
This is carried out efficiently by not using the original features directly. Instead, features
are first mapped to a
query ($Q$) and memory (key and value, $K$ \& $V$) embedding using
linear projections, 
where typically the query and keys are lower dimensional. 
The output for the query
is computed as an attention weighted sum of values $V$, with  the attention weights obtained from
the product of the query $Q$ with 
keys 
$K$. 
In practice, the query here was the word being translated, and the keys and
values are linear projections of the input sequence and the output
sequence generated so far. A location embedding is also added to these
representations in order to incorporate positional information which
is lost in this non-convolutional setup. We refer the readers
to~\cite{vaswani2017attention} and \cite{parmar2018image} for a more
detailed description of the original architecture.

{\noindent \bf \Tx{}:} We now describe our re-purposed Transformer
architecture for the task of video understanding.  Our transformer
unit takes as input the video feature representation and the box
proposal from RPN and maps it into query and memory features.
Our problem setup has a natural
choice for the 
 query ($Q$),
key ($K$) and value ($V$) tensors: the person being classified is the query,
and the
clip around the person is the memory, projected into key and values.
The unit then processes the query and
memory to output an updated query vector. 
The intuition is that the self-attention will add context from other people and objects in the clip to
the query vector, to aid with the subsequent classification.
This unit can be stacked in
multiple heads and layers similar to the original
architecture~\cite{vaswani2017attention}, by concatenating the output
from the multiple heads at a given layer, and using the concatenated
feature as the next query. This updated query is then used to again
attend to context features in the following layer. We show this high-level setup and how it fits into our base network highlighted in green in
Figure~\ref{fig:base_arch}, with each \Tx{} unit denoted as `Tx'. We now explain this unit in detail.

The key
and value features are simply computed as linear projections  of the
original feature map from the trunk, hence each is of shape $T'\times
H'\times W' \times D$.
In practice, we extract
the RoIPool-ed feature for the person box from the center clip, 
and
pass it through a query preprocessor (\QPr{}) 
and a linear layer 
to get
the query feature of size $1\times 1\times D$.  The \QPr{} could directly
average the RoIpool feature across space,
but would lose all spatial layout of the person.  Instead,
we first reduce the
dimensionality by a $1\times 1$ convolution, and then concatenate the cells of the resulting
$7\times 7$ feature map into a vector.  Finally, we reduce the
dimensionality of this feature map using a linear layer to 128D 
(the same as the query and key feature maps).
We refer to this procedure as {\bf
\qprconcat{}} query preprocessing.  We compare this to a \QPr{} that simply averages the
feature spatially, or {\bf \qpravg{}} preprocessing, in Section~\ref{sec:exp:tx_det}.

The remaining architecture essentially follows the Transformer. We use feature $Q^{(r)}$ corresponding to the RPN proposal $r$, for dot-product attention over the $K$ features, normalized by $\sqrt{D}$ (same as~\cite{vaswani2017attention}), and use the result for weighted averaging ($A^{(r)}$) of $V$ features. This operation can be succinctly represented as
\begin{align*}
a_{xyt}^{(r)} = \frac{Q^{(r)}K_{xyt}^T}{\sqrt{D}};
A^{(r)} = \sum_{x,y,t} \left[ \Softmax \left( a^{(r)} \right) \right]_{xyt} V_{xyt}
\end{align*}
We apply a dropout to $A^{(r)}$ and add it to the original query feature. The resulting query is 
passed through a residual branch consisting of a LayerNorm~\cite{ba2016layer} operation, followed by a Feed Forward Network (FFN) implemented as a 2-layer MLP and dropout.
The final feature is passed through one more LayerNorm to get the updated query ($Q''$).
Figure~\ref{fig:base_arch} (Tx unit) illustrates the unit architecture described above, and can be represented as
\begin{align*}
Q^{(r)'} &= \mathrm{LayerNorm} \left( Q^{(r)} + \mathrm{Dropout}\left( A^{(r)} \right) \right) \\
Q^{(r)''} &= \mathrm{LayerNorm} \left( Q^{(r)'} + \mathrm{Dropout} \left( \mathrm{FFN} \left( Q^{(r)'}\right) \right) \right)
\end{align*}

\begin{figure}[t]
\centering
\includegraphics[width=\linewidth]{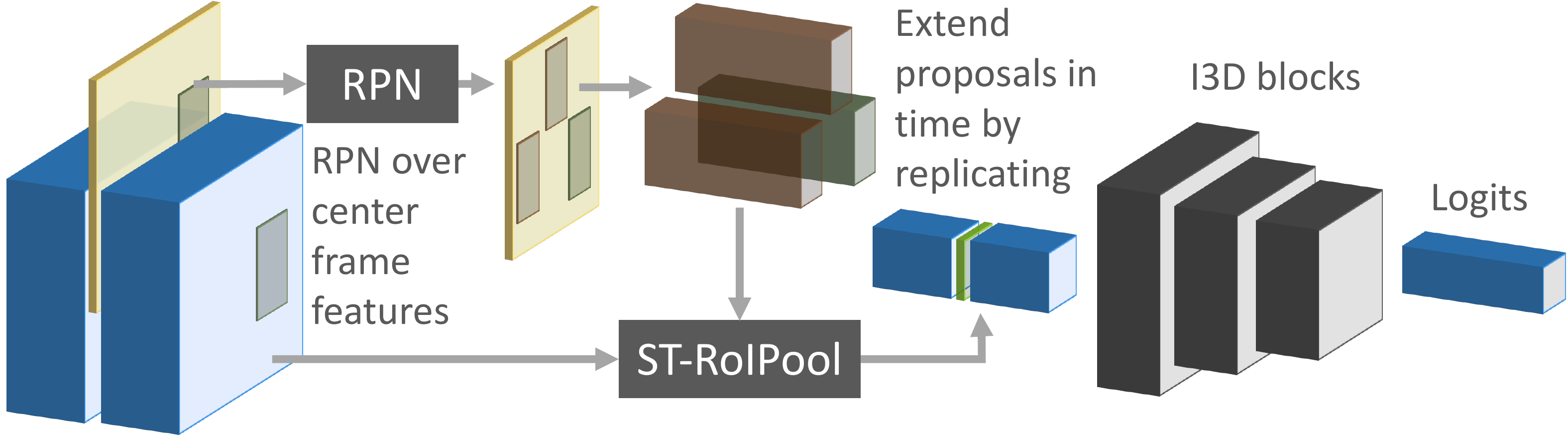}
\caption{
{\bf I3D Head.}
Optionally, we can replace the \Tx{} head with a simpler head that applies the last few I3D blocks to the region features,
as described in Section~\ref{sec:app:i3d_head}.
}\label{fig:i3d_head_arch}
\end{figure}

\subsection{I3D Head}\label{sec:app:i3d_head}

To measure the importance of the context gathered by our \Tx{} head, we also built a 
simpler head architecture that does not extract context.
For this, we extract a feature representation corresponding to the RPN proposal from the feature map using a Spatio-Temporal RoIPool (ST-RoIPool) operation. It's implemented by
first stretching the RP in time by replicating the box to form a tube. Then, we extract a feature representation from feature map at each time point using the corresponding box from the tube using the standard RoIPool operation~\cite{huang2017speed}, similar to previous works~\cite{girdhar2018detecttrack}. The resulting features across time are stacked to get a spatio-temporal feature map corresponding to the tube. 
It is then passed through the layers of the I3D network that were dropped from the trunk (i.e., {\tt Mixed\_5a} to {\tt Mixed\_5c}). The resulting feature map is then passed through linear layers for classification and bounding box regression.
Figure~\ref{fig:i3d_head_arch} illustrates this architecture.

\subsection{Implementation Details}\label{sec:app:impl}
We develop our models in Tensorflow, on top of the TF object detection
API~\cite{huang2017speed}. We use input spatial resolution of
$400\times 400$px and temporal resolution ($T$) of 64.  The RoIPool
used for both the I3D and \Tx{} head generates a $14\times 14$ output,
followed by a max pool to get a $7\times 7$ feature map. Hence, the
I3D head input ends up being $16\times 7\times 7$ in size, while for
\Tx{} we use the $7\times 7$ feature as query and the full $16\times
25\times 25$ trunk feature as the context.  
As also observed in prior work~\cite{vaswani2017attention,parmar2018image},
adding a location embedding in such architectures is very beneficial.
It allows our model to encode \st{} proximity in addition
to visual similarity, a property lost when moving away from traditional convolutional
or memory-based (eg. LSTM) architectures.
For each cell in the trunk feature map, we add explicit location information by constructing
vectors: $[h, w]$ and $[t]$ denoting the spatial and temporal location
of that feature, computed with respect to the size and relative to the center of the feature map. 
We pass each through a 2-layer MLP, and concatenate
the outputs. We then attach the resulting vector to the trunk feature
map along channel dimension.  Since $K, V$
are projections the trunk feature map, and $Q$ is extracted from that
feature via RoIPool, all of these will implicitly contain the location
embedding.
Finally, for classification loss, we use separate logistic losses for each action
class, implemented using sigmoid cross-entropy, since multiple actions can be active
for a given person. For regression,
we use the standard smooth L1 loss.  For the \Tx{} heads, we use
feature dimensionality of $D=128$ and dropout of 0.3. We use a 2-head,
3-layer setup for the \Tx{} units by default, though we ablate other
choices in Section~\ref{sec:exp:tx_ablate}.

\subsection{Training Details}\label{sec:app:train}

{\noindent \bf Pre-training:} 
We initialize most of our models by
pre-training the I3D layers separately on the large,
well-labeled action classification dataset Kinetics-400~\cite{kay2017kinetics} as described in~\cite{carreira2017quo}.
We initialize the remaining layers of our model (eg. RPN, \Tx{} heads etc) from
scratch, fix the running mean and variance statistics of batch norm layers to
the initialization from the pre-trained model,
and then finetune the full model end-to-end.
Note that the only batch norm layers in our model are in the I3D base and head networks;
hence, no new batch statistics need to be estimated when finetuning from the pretrained models.

{\noindent \bf Data Augmentation:} We augment our training data using random flips and crops. We find this was critical, as removing augmentation lead to severe overfitting and a significant drop in performance.
We evaluate the importance of pre-training and data augmentation in Section~\ref{sec:exp:tx_ablate}.

{\noindent \bf SGD Parameters:} The training is done using
synchronized SGD over V100 GPUs with an effective batch size of 30
clips per gradient step. This is typically realized by a per-GPU batch
of 3 clips, and total of 10 replicas. However, since we keep batch
norm fixed for all experiments except for from-scratch experiments,
this batch size can be realized by
splitting the batch over 10, 15 or even 30 replicas for our heavier
models. 
Most of our models are trained for 500K iterations, which
takes about a week on 10 GPUs. We use a learning rate of 0.1 with
cosine learning rate annealing over the 500K iterations, though with a
linear warmup~\cite{goyal2017accurate} from 0.01 to 0.1 for the first
1000 iterations.  For some cases, like models with \Tx{} head and
using ground truth boxes
(Section~\ref{sec:exp:tx_gt}), we stop training early at 300K
iterations as it learns much faster.  The models are trained using
standard loss functions used for object
detection~\cite{huang2017speed}, except for sigmoid cross-entropy for the multi-label classification loss.

 \section{Experiments}\label{sec:exp}

In this section we experimentally evaluate the model on the AVA
benchmark. We start with introducing the dataset and evaluation
protocol in Section~\ref{sec:exp:data}.  Note that the model is
required to carry out two distinct tasks: action {\em localization}
and action {\em classification}. To better understand the challenge of  each independently,
we evaluate each task given perfect
information for the other. In Section~\ref{sec:exp:tx_gt}, we
replace  the RPN proposals with the groundtruth (GT) boxes,   and keep
the remaining architecture as is.  Then in
Section~\ref{sec:exp:tx_det}, we assume perfect classification  by
converting all class labels into a single `active' class label,
reducing the problem into a pure `active person' vs background
detection problem, and evaluate the person
localization performance. Finally we put the lessons from the two together
in Section~\ref{sec:exp:tx_final}. 
We perform all these ablative comparisons on the AVA validation set,
and compare with the state of the art on the test set in Section~\ref{sec:exp:sota}.

\subsection{The AVA Dataset and Evaluation}\label{sec:exp:data}
The Atomic Visual Actions (AVA)
v2.1~\cite{gu2018ava} dataset contains 211K training, 57K
validation and 117K testing clips, taken at 1 FPS from 430 15-minute
movie clips. The center frame in each clip is exhaustively labeled
with all the person bounding boxes, along with one or more of the 80
action classes active for each instance. Following previous
works~\cite{gu2018ava,sun2018arcn}, we report our performance on the
subset of 60 classes that have at least 25 validation examples.  For
comparison with other challenge submissions, we also report the
performance of our final model on the test set, as reported from the
challenge server.  Unless otherwise specified, the evaluation is
performed using frame-level mean average precision (frame-AP) at IOU
threshold of 0.5, as described in~\cite{gu2018ava}.

\subsection{Action classification given GT person boxes}\label{sec:exp:tx_gt}

\begin{table}
\tableSize{}
\setlength{\tabcolsep}{3pt}
\begin{center}
\begin{tabular}{ccccccg}
\toprule
Trunk & Head & \QPr{} & GT Boxes & Params (M) & GFlops & Val mAP \\
\midrule
I3D & I3D & - & & 16.2 & 6.5 & 21.3 \\
I3D & I3D & - & \checkmark & 16.2 & 6.5 & 23.4 \\  
\arrayrulecolor{GrayLine}
\midrule
I3D & Tx & \qpravg{} & & 13.9 & 33.2 & 17.8 \\  
I3D & Tx & \qprconcat{} & & 19.3 & 39.6 & 18.9 \\
I3D & Tx & \qpravg{} & \checkmark & 13.9 & 33.2 & 29.1\\
I3D & Tx & \qprconcat{} & \checkmark & 19.3 & 39.6 & 27.6 \\
\arrayrulecolor{black}
\bottomrule
\end{tabular}
\end{center}
\caption{
{\bf Action classification with GT person boxes.} 
To isolate classification from localization performance,
we evaluate our models when assuming groundtruth box locations are known. 
It can be seen that the \tx{} head has far
stronger  performance than the I3D head when GT boxes are used. All performance
 reported with $R=64$ proposals.
To put the complexity numbers into perspective, a typical video recognition model, 16-frame R(2+1)D network on
Kinetics, is 41 GFlops~\cite{tran2018closer}.
For a sense of random variation, we retrain the basic Tx model (line 5) three times, and get a std deviation of 0.45 (on an mAP of 29.1).
}\label{tab:tx_gt}
\end{table}

In this section we 
assess  how well the head can classify the actions, given the
ground truth bounding boxes provided with the AVA dataset. This will give an upper bound on the action
classification performance of the entire network, 
as the RPN is likely to be less perfect than ground truth.
We start by comparing the I3D head with and without GT
boxes in Table~\ref{tab:tx_gt}.
We use a lower value of $R=64$ for the RPN, in order to reduce the computational expense of these experiments.
It is interesting to note that we only
get a small improvement by using groundtruth (GT) boxes, indicating that
our model is already capable of learning a good representation for
person detection. Next, we 
 replace the I3D head
architecture with the \tx{}, which
leads to a significant 5\% boost for the GT boxes case. 
It is also worth noting that our \tx{} head implementation actually has
2.3M fewer parameters than the I3D head in the LowRes QPr case, dispelling any concerns that
this improvement is simply from additional model capacity. 
The significant drop in performance with and without GT boxes for the  \tx{} is due to only using $R=64$ proposals. As will be seen in subsequent results, this drop is eliminated when the full model with $R=300$ proposals is used.

\subsection{Localization performance (action agnostic)}\label{sec:exp:tx_det}

\begin{table}[t]
\tableSize{}
\setlength\tabcolsep{3pt} 
\begin{center}
\begin{tabular}{cccgc}
\toprule
RoI source & \QPr{} & Head & \multicolumn{2}{c}{Val mAP} \\
& & & IOU@0.5 & IOU@0.75 \\
\midrule
RPN & - & I3D & 92.9 & 77.5 \\
RPN & \qpravg{} & Tx & 77.5 & 43.5 \\
RPN & \qprconcat{} & Tx & 87.7 & 63.3 \\
\arrayrulecolor{black}
\bottomrule
\end{tabular}
\end{center}
\caption{
{\bf Localization performance (action agnostic).}
We perform classification-agnostic evaluation to evaluate the performance of the heads
for person detection. We observe that the I3D head  is superior to
\tx{}-head model, though using the HighRes query transformation (QPr) improves it
significantly. All performance
 reported with $R=64$ proposals.
}\label{tab:tx_det}
\end{table}

Given the strong performance of the \tx{} for the classification task, we look now in detail
 to the localization task.  As described previously, we isolate
the localization performance by merging all classes into a single trivial one. We report performance in Table~\ref{tab:tx_det}, both
with the standard 0.5 IOU threshold, and also with a stricter 0.75
IOU threshold. 

The  I3D head with RPN boxes excels on this task, achieving almost 93\% mAP at 0.5 IOU. The naive implementation of the transformer using a low-resolution query does quite poorly at 77.5\%,   but by adopting the high-resolution query, the gap in performance is considerably reduced (92.9\% to 87.7\%, for the IOU-0.5 metric). The transformer is less accurate for localization and this can be understood by its more global nature; additional research on this problem is warranted. However as we will show next, using the HighRes query we can already achieve a positive trade-off in performance and can leverage the classification gains to obtain a significant overall improvement.

\subsection{Putting things together}\label{sec:exp:tx_final}

\begin{table}[t]
\tableSize{}
\setlength\tabcolsep{2pt} 
\centering
\begin{minipage}[c]{0.5\linewidth}
\resizebox{\linewidth}{!}{
\begin{tabular}{cccHg}
\toprule
Head & \QPr{} & \#proposals & Params (M) & Val mAP \\
\midrule
I3D & - & 64 & 16.2 & 21.3 \\
I3D & - & 300 &  16.2 & 20.5 \\
\arrayrulecolor{GrayLine}
\midrule
Tx & \qprconcat{} & 64 & 19.3 & 18.9 \\
Tx & \qprconcat{} & 300 & 19.3 & 24.4 \\
Tx+I3D & \qprconcat{} & 300 & 23.9 & 24.9 \\
\arrayrulecolor{black}
\bottomrule
\end{tabular}
}
\end{minipage}
\hfill
\resizebox{0.48\linewidth}{!}{
\begin{minipage}[c]{0.6\linewidth}
\vspace{4mm}
\caption{
{\bf Overall  performance.}
Putting the \Tx{} head with \qprconcat{} preprocessing and 300 proposals leads to a  significant improvement
over the I3D head. Using both heads: I3D for regression and Tx for classification performs best.
}\label{tab:tx_final}
\end{minipage}
}
\end{table}

Now we put the transformer head together with the RPN base, and apply the entire network
to the tasks  of detection and classification. We report our findings in
Table~\ref{tab:tx_final}. It can be seen that the \Tx{} head is far superior to the I3D head (24.4 compared to 20.5). An additional boost can be obtained (to 24.9) by using the I3D head for regression and the \Tx{} head for classification -- reflecting their strengths identified in the previous sections -- albeit at a slightly higher (0.1GFlops) computational overhead.

\subsection{Ablation study}\label{sec:exp:tx_ablate}

All our models so far have used class agnostic regression, data augmentation and Kinetics~\cite{kay2017kinetics} pre-training, techniques we observed early on to
be critical for good performance on this task.
We now validate the importance of those
design choices.
We compare the performance using the I3D head network as the baseline in Table~\ref{tab:base_network}. As evident from the table, all three are crucial in getting strong performance. In particular, class agnostic regression is an important
contribution. While typical object detection frameworks~\cite{he2017mask,huang2017speed} learn a separate regression layers for each object category, it does not make sense in our case as the `object' is always a human. Sharing those parameters helps classes with few examples to also learn a good person regressor, leading to an overall boost.
Finally, we note the importance of using a sufficient number of proposals in the RPN. As can be seen 
in
Table~\ref{tab:tx_final}, reducing the number from 
300 to 64 decreases performance significantly
for the  \Tx{} model. The I3D head is less affected. It is interesting because, even for 64, we are using far more proposals than the actual number of people in the frame.

\begin{table}
\tableSize{}
\begin{center}
\begin{tabular}{cHc|ccc}
\toprule
& \begin{tabular}{@{}c@{}} Previous \\ SOTA~\cite{sun2018arcn} \end{tabular} & \begin{tabular}{@{}c@{}}I3D \\ head \end{tabular} & \begin{tabular}{@{}c@{}} Cls-specific \\ bbox-reg \end{tabular} & \begin{tabular}{@{}c@{}} No \\ Data Aug \end{tabular} & \begin{tabular}{@{}c@{}} From \\ Scratch \end{tabular} \\
\midrule
Val mAP & 17.4 & 21.3 & 19.2 &
16.6
& 19.1
\\
\arrayrulecolor{black}
\bottomrule
\end{tabular}
\end{center}
\caption{
{\bf Augmentation, pre-training and class-agnostic regression.} 
We evaluate the importance of certain design choices such as class agnostic box regression, data augmentation and Kinetics pre-training, by reporting the performance when each of those is removed from the model. We use the I3D head model as the baseline. Clearly, removing any leads to a significant drop in performance. All performance
 reported with $R=64$ proposals.
}\label{tab:base_network}
\end{table}

{\bf \noindent Number of heads/layers in \Tx{}:} Our \tx{} architecture is designed to be easily stacked into multiple
heads per layer, and multiple layers, similar to the original
unit~\cite{vaswani2017attention}. We evaluate the effect of changing
the number of heads and layers in Table~\ref{tab:tx_gt_head_layer}. We
find the performance to be largely similar, though tends to get
slightly better with more layers and fewer heads. Hence, we stick with
our default 2-head 3-layer model for all experiments reported in the paper.

{\bf \noindent Swapping out the trunk architecture:}
As we observe in Table~\ref{tab:tx_base_arch}, our model is compatible with different 
trunk architectures. We use I3D for all experiments in the paper given its speed and strong performance.

\begin{table}[t]
\tableSize{}
\begin{center}
\begin{tabular}{c|ccc}
\toprule
 \#layers$\downarrow$ \quad \#heads$\rightarrow$ 
    & 2 & 3 & 6 \\
\midrule
2 & 27.4 & 28.7 & 27.6 \\
3 & 28.5 & 28.8 & 27.7 \\
6 & 29.1 & 28.3 & 26.5 \\
\arrayrulecolor{black}
\bottomrule
\end{tabular}
\end{center}
\caption{
{\bf Ablating the number of heads and layers.} We find fewer heads and more layers tends to give slightly better validation mAP. All performance reported with \tx{} head, when using GT boxes as proposals.
}\label{tab:tx_gt_head_layer}
\end{table}

\begin{table}
\tableSize{}
\begin{center}
\begin{tabular}{cccccg}
\toprule
Trunk & Head & \QPr{} & GT Boxes & Params (M) & Val mAP \\
\midrule
I3D & I3D & - & & 16.2 & 21.3 \\
I3D & I3D & - & \checkmark & 16.2 & 23.4 \\  
I3D & Tx & \qpravg{} & \checkmark & 13.9 & 28.5\\
R3D~\cite{wang2017non} & Tx & \qpravg{} & \checkmark & 17.7 & 26.6 \\
R3D + NL~\cite{wang2017non} & Tx & \qpravg{} & \checkmark & 25.1 & 27.2 \\
\arrayrulecolor{black}
\bottomrule
\end{tabular}
\end{center}
\caption{
{\bf Different trunk architectures.} 
Our model is compatible with different trunk architectures, such as R3D or Non-Local network  proposed in~\cite{wang2017non}. We observed best performance with I3D, so use it for all experiments in the paper.
}\label{tab:tx_base_arch}
\end{table}

\subsection{Comparison with existing state of the art}\label{sec:exp:sota}

\begin{table}[t]
\tableSize{}
\setlength{\tabcolsep}{3pt}
\begin{center}
\resizebox{\linewidth}{!}{
\begin{tabular}{ccccg}
\toprule
Method & Modalities & Architecture & Val mAP & Test mAP \\
\midrule
Single frame~\cite{gu2018ava} & RGB, Flow & R-50, FRCNN & 14.7 & -\\
AVA baseline~\cite{gu2018ava} & RGB, Flow & I3D, FRCNN, R-50 & 15.6 & - \\
ARCN~\cite{sun2018arcn} & RGB, Flow & S3D-G, RN & 17.4 & -  \\
\arrayrulecolor{GrayLine}
\midrule
Fudan University & - & - & - & 17.16 \\
YH Technologies~\cite{yh_ava_submit} & RGB, Flow & P3D, FRCNN & - &  19.60 \\
\begin{tabular}{@{}c@{}} Tsinghua/Megvii~\cite{tsinghua_ava} \end{tabular}
  & RGB, Flow &
  \begin{tabular}{@{}c@{}} I3D, FRCNN, NL, TSN, \\ C2D, P3D, C3D, FPN  \end{tabular}
       & - & 21.08 \\
\arrayrulecolor{GrayLine}
\midrule
Ours (Tx-only head) & RGB & I3D, Tx & 24.4 &  24.30 \\  
Ours (Tx+I3D head) & RGB & I3D, Tx & 24.9 & 24.60 \\  
Ours (Tx+I3D+96f) & RGB & I3D, Tx & {\bf 25.0} & {\bf 24.93} \\  
\arrayrulecolor{black}
\bottomrule
\end{tabular}
}  
\end{center}
\caption{
{\bf Comparison with previous state of the art and challenge submissions.}
Our model outperforms the previous state of the art by $>7.5$\% on the validation set, and the CVPR'18 challenge winner by $>3.5$\% on the test set. We do so while only using a single model (no ensembles), running on raw RGB frames as input. This is in contrast to the various previous methods listed here, which use various modalities and ensembles of multiple architectures.
The model abbreviations used here refer to the following. R-50: ResNet-50~\cite{He_16}, I3D: Inflated 3D convolutions~\cite{carreira2017quo}, S3D(+G): Separable 3D convolutions (with gating)~\cite{xie2017rethinking}, FRCNN: Faster R-CNN~\cite{ren2015faster}, NL: Non-local networks~\cite{wang2017non}, P3D: Pseudo-3D convolutions~\cite{qiu2017learning}, C2D~\cite{tran2018closer}, C3D~\cite{tran2018closer}, TSN: Temporal Segment Networks~\cite{WangL_16a} RN: Relation Nets~\cite{santoro2017simple}, Tx: Transformer~\cite{vaswani2017attention,parmar2018image} and FPN: Feature Pyramid Networks~\cite{lin2017feature}. Some of the submissions also attempted to use other modalities like audio, but got lower performance. Here we compare with their
best reported numbers. 
}\label{tab:sota}
\end{table}

Finally, we compare our models to the previous state of the art on the test set in Table~\ref{tab:sota}.
We find the Tx+I3D head obtains the best performance, and simply adding temporal context at test time (96 frames compared to 64 frames at training) leads to a further improvement. We outperform the previous state of the art by more than 7.5\% absolute points on validation set, and the CVPR 2018 challenge winner by more than 3.5\%.  It is also worth noting that our approach is much simpler than most previously proposed approaches, especially the challenge submissions that are ensembles of multiple complex models. Moreover, we obtain this performance only using raw RGB frames as input, while prior works 
use RGB, flow, and in some cases audio as well.
 \begin{figure*}[t]
\centering
\centering\begin{tabularx}{\textwidth}{XXX@{ }XXX@{ }XXXXX}
Frame & Tx-A & Tx-B & Frame & Tx-A & Tx-B &
\multicolumn{5}{>{\hsize=\dimexpr5\hsize+5\tabcolsep+\arrayrulewidth\relax}Y}{Attention} \\
\end{tabularx}

\includegraphics[width=\linewidth]{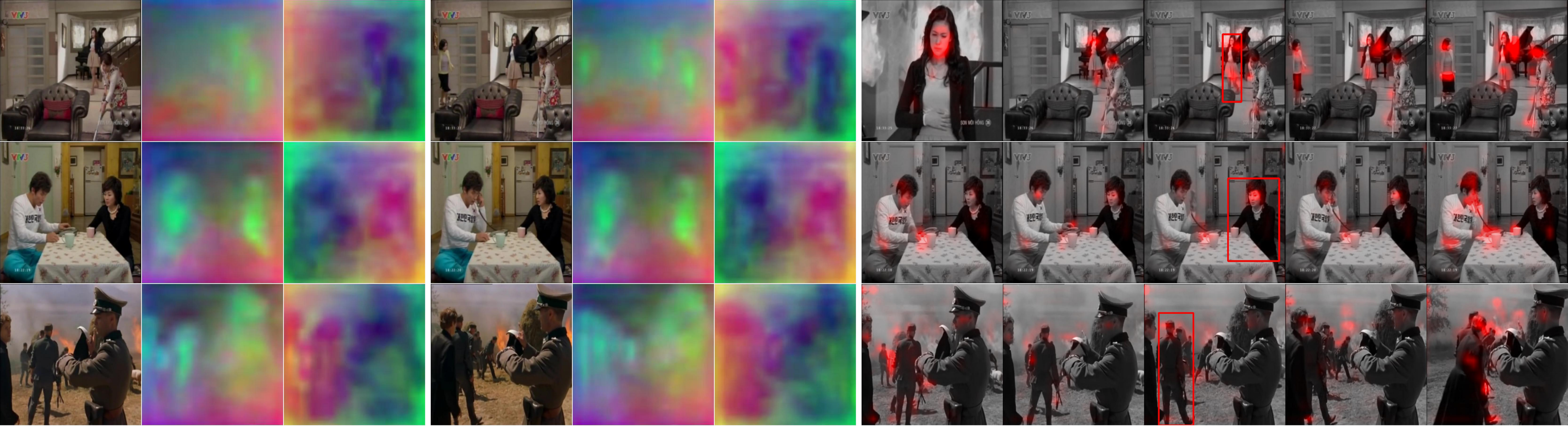}
\caption{
{\bf Embedding and attention.}
For two frames, we show their `key' embeddings as color-coded 3D PCA projection for two of the six heads in our 2-head 3-layer Tx head.
It is interesting to note that one of these heads learns to track people semantically (Tx-A: all upper bodies are similar color -- green), while the other is instance specific (Tx-B: each person is different color -- blue, pink and purple).
In the following columns we show by the average softmax attention corresponding to the person in the red box for all heads in the last Tx layer. Our model learns to hone in on faces, hands and objects being interacted with, as these are most discriminative for recognizing actions. 
}\label{fig:analysis:embedding}
\end{figure*}

\begin{figure*}[t]
\centering
\begin{minipage}[c]{0.66\linewidth}
\resizebox{\linewidth}{!}{
{\footnotesize (a)} \raisebox{-.5\height}{\includegraphics[width=\linewidth]{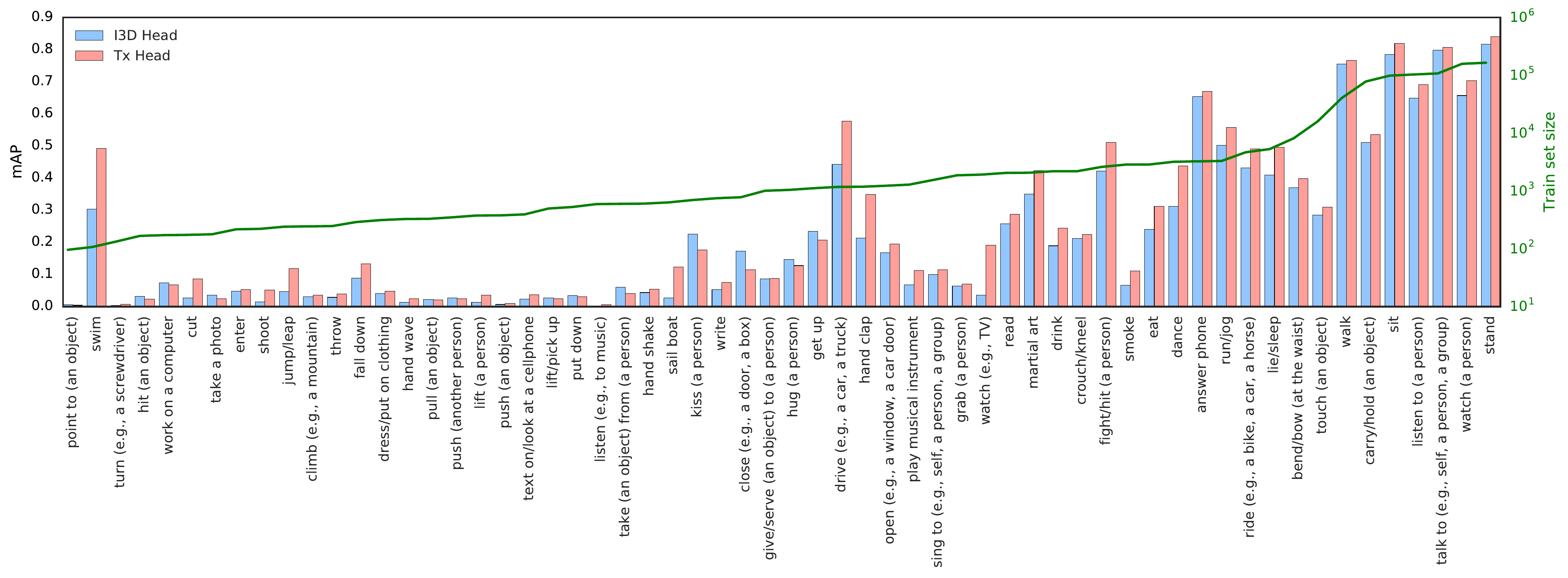}}
}
\end{minipage}\hfill
\begin{minipage}[c]{0.33\linewidth}
{\footnotesize (b)} \raisebox{-.5\height}{\includegraphics[width=\linewidth]{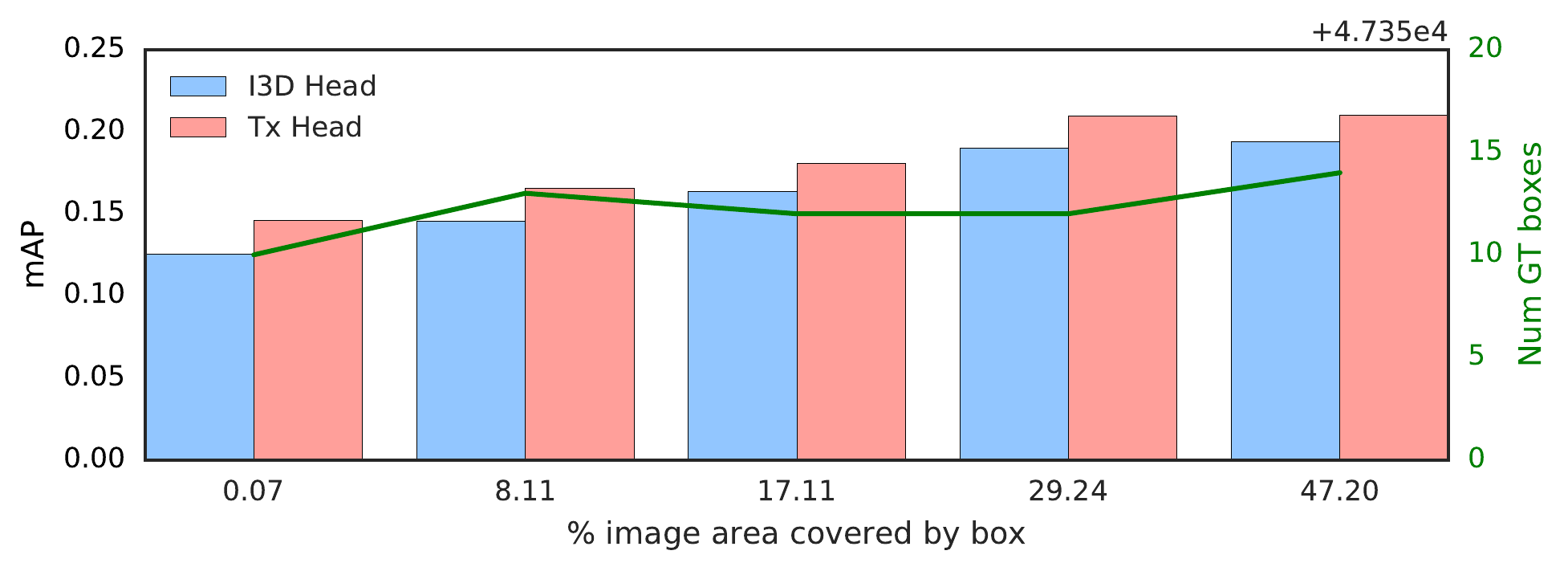}}\hfill
{\footnotesize (c)} \raisebox{-.5\height}{\includegraphics[width=\linewidth]{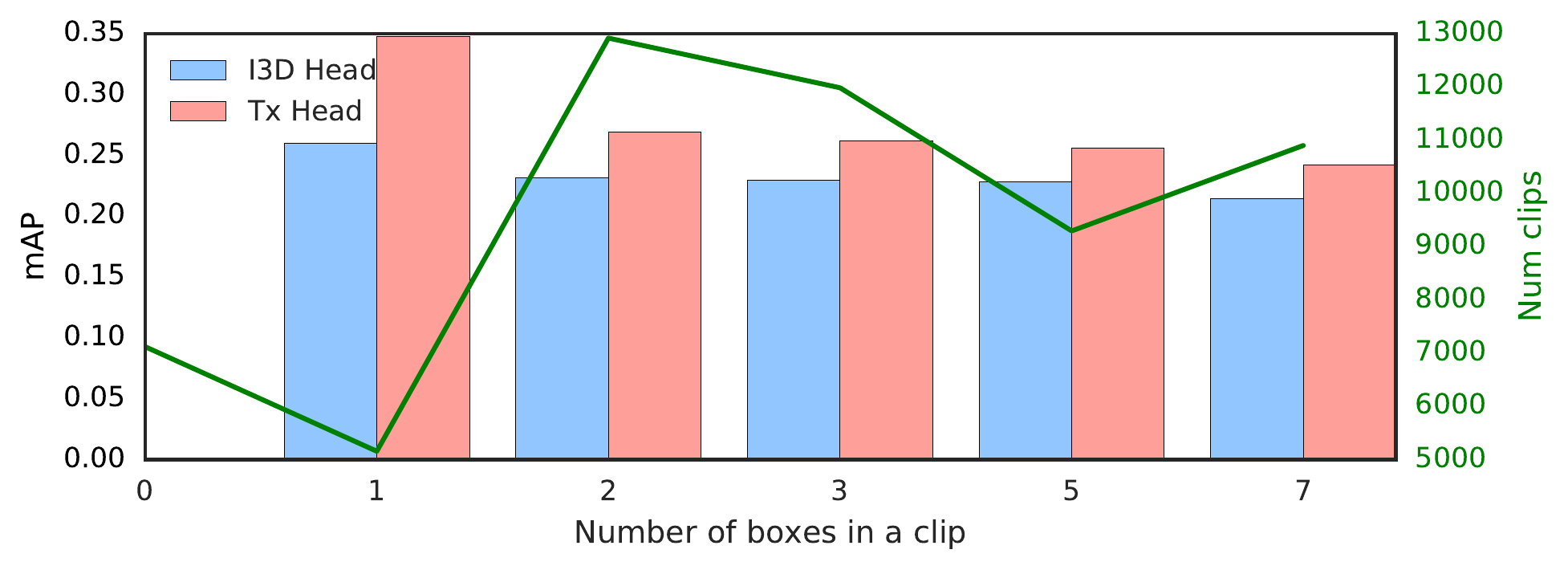}}
\end{minipage}
\caption{
{\bf Performance by (a) class, (b) box area and (c) count.}
While overall trend suggests a positive correlation of performance with train-set size (green line), there do exist interesting anomalies such as `smoking', `eating' etc, which are still hard to recognize despite substantial training data. 
In (b) and (c) the green line denotes the validation subset size. We observe the performance largely improves as the person box size increases, and as number
of boxes decreases. {\em Axis labels best viewed zoomed in on screen.}
}\label{fig:analysis:perf_by_area}\label{fig:analysis:perf_by_box_count-area}\label{fig:analysis:per_class}
\end{figure*}

\section{Analysis}
\label{sec:analysis}

We now analyze the \tx{} model. Apart from obtaining superior performance, this model is also more interpretable by explicitly encoding bottom up attention. We start by visualizing the key/value embeddings and attention maps learned by the model. 
Next we analyze the performance vis-a-vis specific classes, person sizes and counts; and finally visualize common failure modes. 

{\bf \noindent Learned embeddings and attention:}
We visualize the 128D `key' embeddings and attention maps in Figure~\ref{fig:analysis:embedding}. We visualize the embeddings by color-coding a 3D PCA projection. We show two heads out of the six in our 2-head 3-layer \Tx{} model. For attention maps, we visualize the average softmax attention over the 2 heads in the last layer of our Tx head. It is interesting to note that our model learns to track the people over the clips, as shown from the embeddings where all `person' pixels are same color. Moreover, for the first head all humans have the same color, suggesting a {\em semantic} embedding, while the other has different, suggesting an {\em instance-level} embedding.
Similarly, the softmax attention maps learn to attend and track faces, hands and other parts of the person of interest as well as the other people in the scene. It also tends to attend to objects the person interacts with, like the vaccum cleaner and coffee mugs.
This makes sense as many actions in AVA such as talking, listening, hold an object etc.\ require focusing the faces, hands of people and objects to deduce. 
A video visualization of the embeddings, attention maps and predictions are provided in the supplementary video~\cite{supp_video}.

\begin{figure}[t]
\centering
\includegraphics[width=\linewidth]{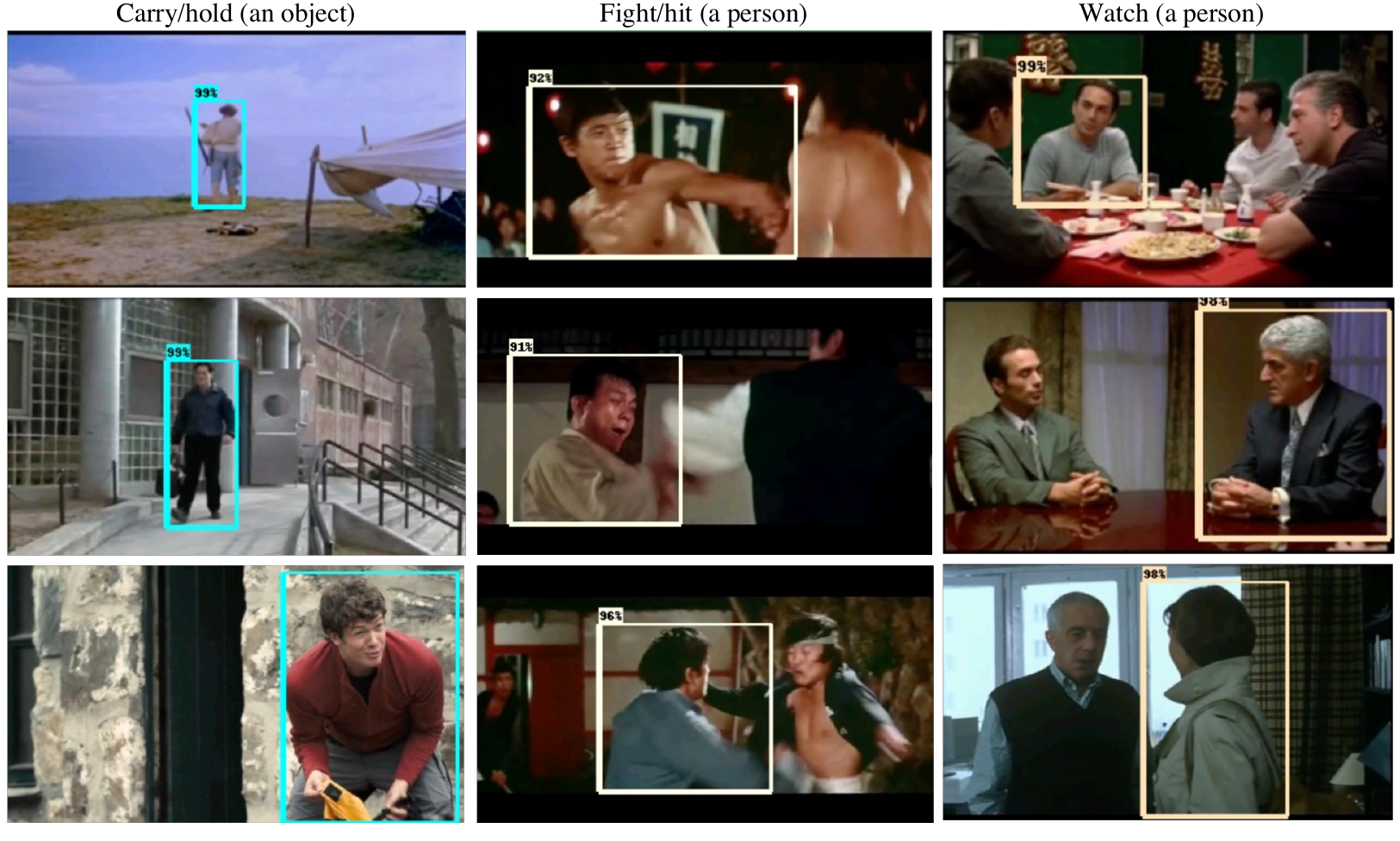}
\caption{
{\bf Top predictions.}
Example top predictions for some of the classes using our model.
Note that context, such as other people or objects being interacted with, is often helpful for the classifying actions like `watching a person', `holding an object' and so on. Capturing context is a strength of our model.
}\label{fig:analysis:true_p}
\end{figure}

\begin{figure}[t]
\centering
\includegraphics[width=\linewidth]{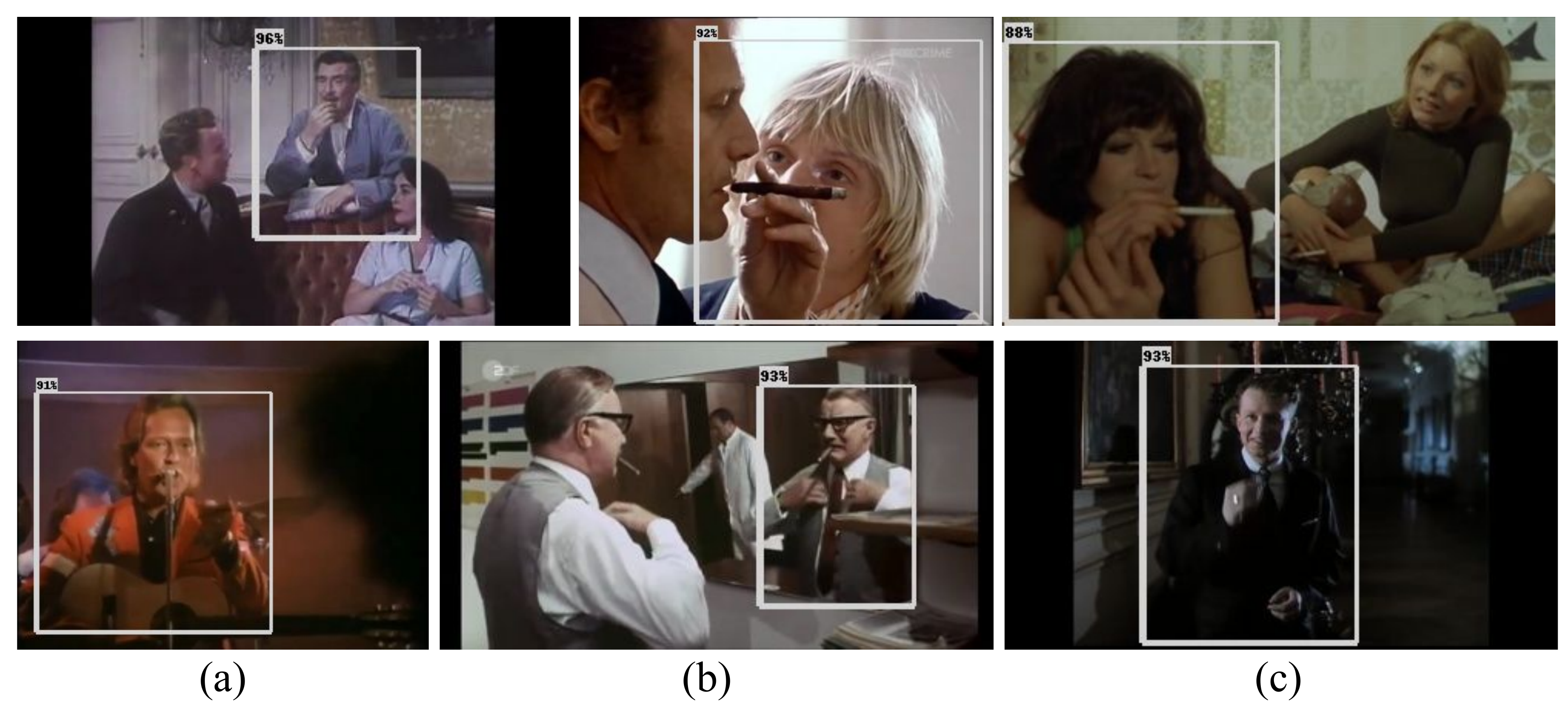}
\caption{
{\bf Misclassified videos.}
Videos from the `smoking' class that obtains low performance even with large amount of training data.
Failure modes include, (a) {\em Similar action/interaction:} In the first clip, the person has his hand on his mouth, similar to a smoker; and in the second, the mic looks like a cigarette; (b) {\em Identity:} There are multiple people (or reflections) and the action is not being assigned to the correct person; (c) {\em Temporal position:} The dataset expects the action to be occurring in the key frame, in these examples the action has either finished or not started by the key frame.
}\label{fig:analysis:false_p}
\end{figure}

{\bf \noindent Breaking down the performance:}
We now break down the performance of our model into certain bins. 
We start by evaluating the performance per class in Figure~\ref{fig:analysis:per_class} (a). We sort the performance according the increasing amounts of training data, shown in green. While there is some correlation between the training data size and performance, we note that there exist many classes with enough data but poor performance, like smoking. We note that we get some of the largest improvement in classes such as sailing boat, watching TV etc, which would benefit from our \tx{} model attending to the context of the person.
Next, we evaluate the performance with respect to the size of the person in the clip, defined by the percentage area occupied by the GT box, in Figure~\ref{fig:analysis:perf_by_box_count-area} (b). For this, we split the validation set into bins, keeping predictions and GT within certain size limits. We find the size thresholds by sorting all the GT boxes and splitting into similar sized bins, hence ensuring similar `random' performance for each bin. We find performance generally increases with bigger boxes, presumably because it becomes progressively easier to see what the person is doing up close.
Finally, we evaluate the performance with respect to the number of GT boxes labeled in a clip in Figure~\ref{fig:analysis:perf_by_box_count-area} (c). We find decreasing performance as we add more people in a scene.

{\bf \noindent Qualitative Results:}
We visualize some successes of our model in Figure~\ref{fig:analysis:true_p}.
Our model is able to exploit the context to recognize actions such as `watching a person', which are inherently hard when just looking at the actor.
Finally, we analyze some common failure modes of our best model in Figure~\ref{fig:analysis:false_p}. The columns show some common failure modes like (a) similar action/interaction, (b) identity and (c) temporal position.
A similar visualization of top predictions on the validation set for each class, sorted by confidence, is provided in~\cite{preds}.

 \section{Conclusion}

We have shown that the \Tx{} network is able to learn spatio-temporal context from other human actions
and objects in a video clip to localize and classify human actions. The resulting embeddings and attention
maps (learned indirectly as part of the supervised action training) have 
a semantic meaning. 
The network exceeds the state-of-the-art on the AVA dataset by a significant margin.
It is worth noting that previous state-of-the-art networks
have used a motion/flow stream in addition to RGB~\cite{carreira2017quo,xie2017rethinking}, so adding
flow as input is likely to  boost performance also for the \Tx{} network.
Nevertheless, 
performance is far from perfect, and we have suggested
several avenues for improvement and investigation. %

{\small
{\bf \noindent Acknowledgements:}
We would like to thank V.\ Patraucean, R.\ Arandjelovi\'{c}, J.-B.\ Alayrac, A.\ Arnab, M.\ Malinowski and C.\ McCoy for helpful discussions.
} 
{\small
\bibliographystyle{ieee}
\bibliography{refs}
}

\end{document}